\documentclass[10pt,twocolumn,letterpaper]{article}

\usepackage{iccv}
\usepackage{times}
\usepackage{epsfig}
\usepackage{graphicx}
\usepackage{amsmath}
\usepackage{amssymb}
\usepackage{algorithm}
\usepackage{algorithmic}
\usepackage{mathrsfs}
\usepackage{multirow}

\usepackage[]{animate}

\usepackage{graphicx}
\usepackage{animate}

\usepackage{float}

\usepackage{multirow}
\usepackage{rotating}
\usepackage{wrapfig}
\usepackage{lineno}
\usepackage{epsfig}
\usepackage{amsmath,amssymb}
\usepackage{color}
\usepackage{booktabs}       
\usepackage{amsfonts,amsmath}       
\usepackage{nicefrac}       
\usepackage{microtype} 
\usepackage{float}
\usepackage{amsmath}

\usepackage{enumerate}
\usepackage{enumitem}   

\usepackage{cancel}

\usepackage[pagebackref=true,breaklinks=true,letterpaper=true,colorlinks,bookmarks=false]{hyperref}

\iccvfinalcopy 

\def\iccvPaperID{9313} 

\ificcvfinal\fi
\begin{document}

\title{Adversarial Unsupervised Domain Adaptation with Conditional and Label Shift: Infer, Align and Iterate}

\author{Xiaofeng Liu$^{1}$, Zhenhua Guo$^{2}$, Site Li$^{3}$, Fangxu Xing$^{1}$, Jane You$^{4}$, C.-C. Jay Kuo$^{5}$,\\ Georges El Fakhri$^{1}$, Jonghye Woo$^{1}$\\~\\

$^{1}$Gorden Center for Medical Imaging, Massachusetts General Hospital and Harvard Medical School,\\ Boston, MA, USA. \\
$^{2}$Alibaba Group, Hangzhou, Zhejiang, China. \\
$^{3}$Carnegie Mellon University, Pittsburgh, PA, USA. \\
$^{4}$Dept. of Computing, The Hong Kong Polytechnic University, Hong Kong.\\
$^{5}$Dept. of ECE, University of Southern California, Los Angeles, CA, USA.\\
{\tt\small liuxiaofengcmu@gmail.com}
}

\maketitle

\begin{abstract}

In this work, we propose an adversarial unsupervised domain adaptation (UDA) method under inherent conditional and label shifts, in which we aim to align the distributions w.r.t. both $p(x|y)$ and $p(y)$. Since labels are inaccessible in a target domain, conventional adversarial UDA methods assume that $p(y)$ is invariant across domains and rely on aligning $p(x)$ as an alternative to the $p(x|y)$ alignment. To address this, we provide a thorough theoretical and empirical analysis of the conventional adversarial UDA methods under both conditional and label shifts, and propose a novel and practical alternative optimization scheme for adversarial UDA. Specifically, we infer the marginal $p(y)$ and align $p(x|y)$ iteratively at the training stage, and precisely align the posterior $p(y|x)$ at the testing stage. Our experimental results demonstrate its effectiveness on both classification and segmentation UDA and partial UDA.

\end{abstract}

\section{Introduction} 

Deep learning methods are highly reliant on the large volume of labeled training datasets and the independent and identically distributed (i.i.d.) assumption of the training and testing data \cite{che2021deep,liu2021Generalization}. The real world implementation scenarios, however, can be significantly diverse, and it can be costly to label datasets in every target environment \cite{cordts2016cityscapes,patel2015visual}. To address this, unsupervised domain adaptation (UDA) can be used to transfer knowledge learned from a labeled source domain to different unlabeled target domains \cite{ganin2016domain,liu2021adapting,liu2021generative}.


\begin{figure}[t]
\begin{center} 
\includegraphics[width=1\linewidth]{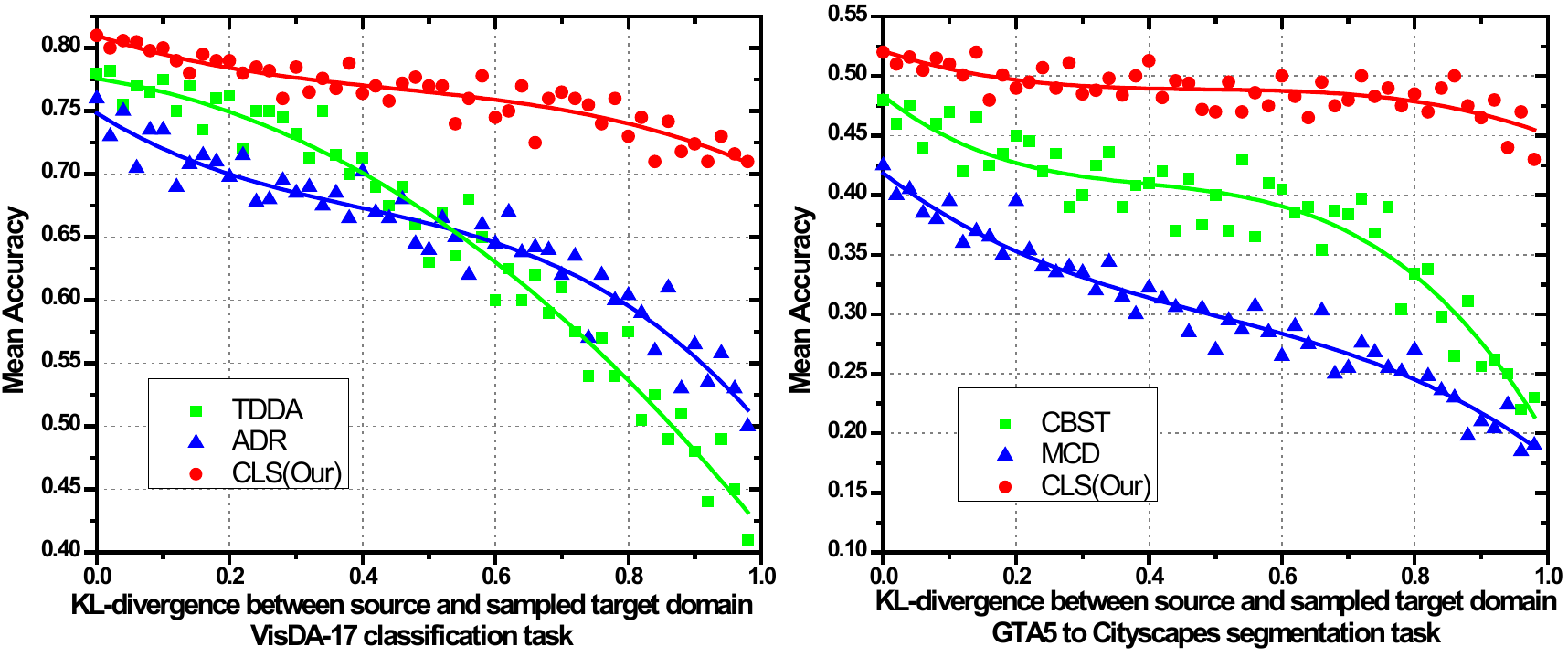} 
\end{center}
\caption{Plots of the UDA performance with different levels of label shift using $KL(p_s(y),p_t(y))$ as a metric. Our CLS is more robust to label shift than conventional adversarial training (TDDA), maximum discrepancy (MCD), dropout (ADR), and self-training (CBST).} 
\label{fig:b}
\end{figure}

One of the predominant streams in UDA makes use of the adversarial training \cite{ganin2014unsupervised,liu2019feature,liu2021mutualpami,liu2021mutualpr}, which hinges on a discriminator to enforce $p_s(f(x))=p_t(f(x))$, where $f (\cdot)$ is a feature extractor. It results in a domain invariant representation under the $covariate$ $shift$ assumption \cite{moreno2012unifying}. However, the $covariate$ $shift$ assumption is not realistic in lieu of $conditional$ $shift$ w.r.t. $p(x|y)$, since each class may have its own appearance shift protocol. For example, a street lamp is sparkly in the night, while a pedestrian is shrouded in darkness.

Essentially, we would expect the fine-grained class-wise alignment w.r.t. $p(x|y)$, while $p(y)$ in a target domain is inaccessible in UDA. Assuming that there is no concept shift (i.e., $ p_s(y|f(x))=p_t(y|f(x))$) and label shift (i.e., $p_s(y)=p_t(y))$), and given the Bayes' theorem, $p(f(x)|y)=\frac{p(y|f(x))p(f(x))}{p(y)}$, adversarial UDA can align $p(f(x)|y)$ if $p_s(f(x))=p_t(f(x))$. The label shift, $i.e.,$ different class proportions, however, is quite common \cite{lipton2018detecting}, e.g., a motorbike is more common in Taipei than in Tokyo.


The adversarial learning based UDA with this unrealistic assumption has been outperformed by self-training \cite{zou2019confidence}, dropout \cite{saito2017adversarial}, and moment matching methods \cite{pan2019transferrable} in most of the benchmarks.

Either the $conditional$ $shift$ \cite{long2018conditional} or $label$ $shift$ \cite{chan2005word,kouw2018introduction} has a long research history. However, it is ill-posed to only consider one of them in UDA without $p_t(y)$ \cite{zhang2013domain,kouw2018introduction}. \cite{zhang2013domain,gong2016domain,combes2020domain} take both the $conditional$ and $label$ shifts into account from a causal interpretation view. However, its linearity assumption of $p_s(x|y)$ and $p_t(x|y)$ does not necessarily hold. Unfortunately, there are few practical methods available for real-world applications. Moreover, the theoretical analysis and methodology under the conditional and label shift assumptions in recently flourished adversarial learning-based UDA warrant a specific discussion as well.


In this work, we theoretically analyze the inequilibrium of conventional adversarial UDA methods under the different shifts. The discrepancy between the marginal distributions $p_s(y)$ and $p_t(y)$ is measured via the $KL$-divergence as the semi-supervised learning with the selective bias problem \cite{zadrozny2004learning}. The impact of a label shift is empirically illustrated in Fig. \ref{fig:b}. 
     
\begin{figure}[t]
\begin{center}
\includegraphics[width=1\linewidth]{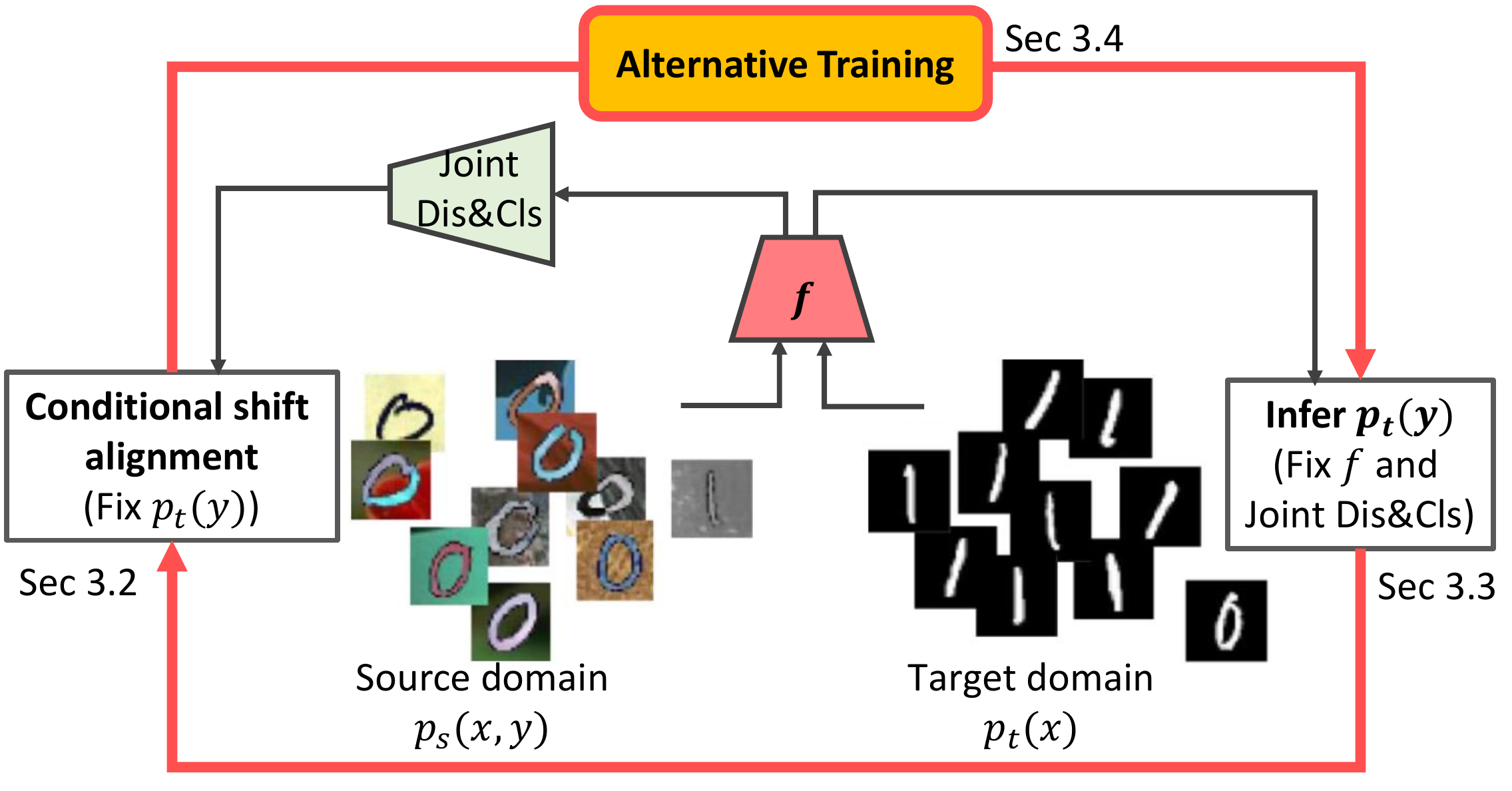}
\end{center} 
\caption{Illustration of our alternative training scheme for conditional and label shifts: \textbf{infer} $p(y)$, \textbf{align} $p(f(x)|y)$, and \textbf{iterate}.} 
\label{fig:c}
\end{figure}

We believe that the conditional and label shift hypothesis is a better and minimal assumption in many real-world scenarios. We first assume the target label distribution $p_t(y)$ can be estimated, and further propose to align the conditional shift between $p_s(x|y)$ and $p_t(x|y)$. To make the discriminator aware of the class label $y$, we adopt the joint discriminator and classifier parameterization and give a theoretical analysis from a domain adaptation theory perspective. A class-level balancing parameter $w.r.t.$ the estimated $p_t(y)$ can be simply applied to adversarial UDA methods, following a plug-and-play manner. Since we are quantifying the transferability of each category instead of each example \cite{cao2019learning}, it is more promising for large-scale tasks, especially for the segmentation with pixel-wise classification. 

Then, we can estimate $p_t(y)$, assuming that $p_s(x|y)$ and $p_t(x|y)$ are aligned using feature-level mean matching, which is incorporated into an alternative optimization framework. The precise target label distribution and a decent classifier are expected after sufficient round-based iterations. At last, we use these two parts to align the posterior $p_t(y|x)$, i.e., target classifier.

   
Our contributions are summarized as follows:

\noindent$\bullet$ We explore both the conditional and label shifts in adversarial UDA, and give a thorough theoretical and empirical analysis of the conventional UDA methods under this assumption. 
   
\noindent$\bullet$ An alternative optimization scheme is proposed to align the conditional and label distributions at the training stage, following the infer, align, and iterate scheme. Finally, we align the posterior $p_t(y|x)$ at the testing stage.
   
\noindent$\bullet$ We propose a practical and scalable method to align the conditional shift with the class-level balancing parameter.

   
We extensively evaluate our method on both popular UDA classification and semantic segmentation benchmarks. It can be naturally generalized to partial UDA.

\section{Related Work}

\textbf{Unsupervised domain adaptation} is aimed at transferring knowledge learned from a labeled source domain to an unlabeled target domain \cite{kouw2018introduction}. The solutions can be mainly classified into statistic moment matching (e.g., maximum mean discrepancy (MMD) \cite{long2018conditional}), domain style transfer \cite{sankaranarayanan2018generate}, self-training \cite{zou2019confidence,liu2021energy,liu2021generative}, and feature-level adversarial learning \cite{ganin2016domain,he2020classification,he2020image2audio,liu2018data}. As one of the mainstreams, adversarial learning based UDA proposes to encourage the features $f(x)$ from different domains indistinguishable. Conditional and label shifts, however, are ubiquitous in real-world tasks. At present, the moment matching and self-training usually dominate the top place in classification and segmentation tasks, respectively.

\begin{figure}[t]
\begin{center}
\includegraphics[width=0.8\linewidth]{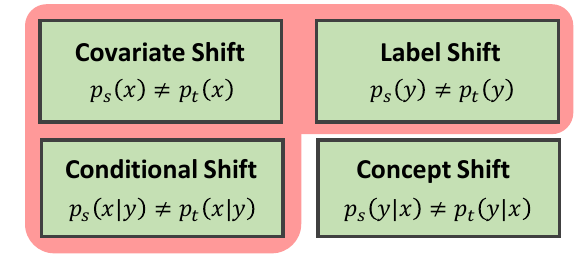}
\end{center} 
\caption{A summary of the possible shifts. The red area indicates the common shifts in UDA. $p(x)$ can be aligned, if $p(x|y)$ is aligned with the law of total probability \cite{zhang2013domain}.}
\label{fig:d} 
\end{figure}

\textbf{Domain shifts} \cite{kouw2018introduction} can be categorized into four classes based on the shift content, as summarized in Fig. \ref{fig:d}. Conventionally, each shift is studied independently, by assuming that the other shift conditions are invariant across domains \cite{kouw2018introduction}.
   
The conditional shift \cite{long2018conditional} is more realistic than the covariate shift assumption \cite{kouw2018introduction}. However, without $p_t(y)$, estimation of $p_t(x|y)$ is, in general, ill-posed \cite{zhang2013domain}. 
   
The label shift \cite{chan2005word} (a.k.a. the target shift) occurs when the proportion of each class is different between source and target domains. \cite{lipton2018detecting} propose a test distribution estimator to detect the label shift. \cite{azizzadenesheli2019regularized} introduce a regularization approach to correct the label shift. \cite{chen2018re} consider the label shift in an optimal transportation-based UDA task, which is also related to the class imbalance problem in the MMD framework. \cite{wu2019domain} propose an asymmetrically-relaxed alignment in adversarial UDA. These methods, however, assume that there is no conditional shift.
   
Moreover, in partial UDA \cite{cao2019learning}, classes in a target domain are a subset of those in a source domain. It can be regarded as a special case of the label shift in that some classes in the target domain are zero. Instead of calculating the transferability of each source example, which can be each pixel in the segmentation task \cite{cao2019learning}, we directly estimate the probability of each class and demonstrate its generality and scalability for a novel segmentation partial UDA task. In a related development, the open set adaptation \cite{panareda2017open} also has new target categories and treats them as unknown, which is different from the above settings.
   
The concept shift \cite{kouw2018introduction} can arise, when classifying a tomato as a vegetable or fruit in different countries. It is usually not a common problem in popular object classification or semantic segmentation tasks.

\cite{zhang2013domain,gong2016domain,combes2020domain} consider the conditional and target shifts in a casual system ($y\rightarrow{x}$) with a somewhat unrealistic linearity assumption. Unfortunately, they are not aligned with the modern UDA methods and sensitive to the scale of the dataset. To the best of our knowledge, this is the first attempt at considering both the conditional and label shifts in adversarial UDA.

 \section{Methodology} 

In UDA, given a source domain $p_s(x,y)$ and a target domain $p_t(x,y)$, a labeled set $\mathcal{D_{S}}$ is drawn $i.i.d.$ from $p_s(x,y)$, and an unlabeled set $\mathcal{D_{T}}$ is drawn $i.i.d.$ from the marginal distribution $p_t(x)$. The goal of UDA is to build a good classifier in the target domain, with the training on $\mathcal{D_{S}}$ and $\mathcal{D_{T}}$. $\mathcal{Y}=\left\{1,2,\dots,c\right\}$ is the set of the class label. Adversarial UDA \cite{ganin2016domain,tzeng2017adversarial} is motivated by the following theorem \cite{kouw2018introduction}: 

\vspace{+5pt}
\noindent\textbf{Theorem 1}
For a hypothesis $h$
{\begin{align}
&\mathcal{L}_t(h)\leq \mathcal{L}_s(h)+d[p_s(x),p_t(x)]+  \label{e:1}\\
&{\rm{min}}[\mathbb{E}_{x\sim p_s}|p_s(y|x)-p_t(y|x)|,\mathbb{E}_{x\sim p_t}|p_s(y|x)-p_t(y|x)|],\nonumber
\end{align}}where $\mathcal{L}_s(h)$ and $\mathcal{L}_t(h)$ denote the expected loss with hypothesis $h$ in the source and target domains, respectively, $d[\cdot]$ is equivalent to the Jensen–Shannon divergence in conventional adversarial UDA \cite{salimans2016improved}.

Based on Theorem \textcolor{red}{1} and assuming that performing feature transform on $x$ will not increase the values of $\mathcal{L}_t(h)$ and last terms of Eq. \ref{e:1}, adversarial UDAs \cite{ganin2016domain,tzeng2017adversarial} apply a feature extractor $f(\cdot)$ onto $x$, hoping to obtain a feature representation $f(x)\in\mathbb{R}^K$ that has a lower value of $d[p_s(f(x)),p_t(f(x))]$. To this end, $f(\cdot)$ is trained to make the feature distributions of the two domains indistinguishable, by using a domain discriminator $Dis:\mathbb{R}^K\rightarrow(0,1)$, and training a classifier $Cls$ to correctly classify the source data. $Cls:\mathbb{R}^K\times\mathcal{Y}\rightarrow(0,1)$ outputs the probability of an extracted feature $f(x)$ being a class $y$ among $c$ categories, i.e., $ C(f(x),y)=p(y|f(x);Cls)$. The objective of different modules can be {\begin{align}
   &~_{Cls}^{\rm{max}}~~^\mathbb{~~E}_{x\sim p_s} {\rm log} {{C}}(f(x),y) \label{e:2}\\ 
   &~_{Dis}^{\rm{max}}~~^\mathbb{~~E}_{x\sim p_s} {\rm log} (1-Dis(f(x))+^\mathbb{~~E}_{x\sim p_t} {\rm log}Dis(f(x)) \label{e:3}\\
   &~_{~f}^{\rm{max}}~^\mathbb{~~E}_{x\sim p_s} {\rm log} {{C}}(f(x),y)+\lambda^\mathbb{~~E}_{x\sim p_t} {\rm log} (1-Dis(f(x)), \label{e:4}
\end{align}}where $\lambda\in\mathbb{R}^+$ balances between the classification and adversarial loss. Here, we follow the conventional adversarial UDA methods that make use of the easily implemented three $max$ \cite{tran2019gotta,salimans2016improved}. Note that {maximizing} $~^\mathbb{~~E}_{x\sim p_t} {\rm log}Dis(f(x))$ for $Dis$ in Eq \ref{e:3}, while {maximizing} $~^\mathbb{~~E}_{x\sim p_t} {\rm log}(1-Dis(f(x)))$ for $f$ in Eq \ref{e:4} has made the $minmax$ adversarial game. We omit the subscript $y$ for simplicity, since only the source domain has $y$ in UDA. 

\subsection{Motivation} 

We first analyze how the $label$ $shift$ affects the adversarial UDA, when enforcing the the distribution of feature representation $f(x)$ to be invariant across domains. We start the analysis under the label shift only assumption in adversarial learning, followed by taking a step to introduce the conditional shift. 

\vspace{+5pt}
\noindent\textbf{Theorem 2} Given $p_s(x|y)=p_t(x|y)$ and $p_s(y)\neq p_t(y)$, if a decent $f(\cdot)$ makes $p_s(f(x))=p_t(f(x))$, then we have $p_s(y|f(x))\neq p_t(y|f(x))$, i.e., the concept shift of $f(x)$, and $\exists~ p_s(y=i|f(x))=p_s(y)$, $p_t(y=i|f(x))=p_t(y=i)$.\label{t:2} 
\vspace{+5pt}

From Theorem \textcolor{red}{2}, under the label shift, aligning $p_s(f(x))$ with $p_t(f(x))$ results in the posterior w.r.t. $f(x)$ inconsistent in different domains, i.e., $concept$ $shift$ is induced. We can also deduce that enforcing $f(x)$ to be domain invariant may mix the feature of all classes in the feature space, since $p_s(y=i|f(x))$ or $p_t(y=i|f(x))$ will be a constant for any $x$. Let us make a specific example that all samples $x$ are mapped onto a consistent point $f_0$ in the feature manifold of $f(x)$; we have $p_s(f(x)=f_0)=p_t(f(x)=f_0)=1$. As a result, $f(x)$ is domain invariant as expected, but the classifier will completely lose the discriminative ability. Actually, strictly enforcing $p_s(f(x))=p_t(f(x))$ in this case will make the classifier less disciminative, i.e., increasing the target loss $\mathcal{L}_t(h)$ or its upper bound $\mathcal{L}_s(h)$ in Theorem \textcolor{red}{1} \cite{kouw2018introduction}. 

Furthermore, under the $joint$ $conditional$ $and$ $label$ $shift$ assumption, there is also a conflict between minimal domain discrepancy and superior classification performance. Following the law of total probability, the optimization objective of adversarial UDA to align $p_s(f(x))$ with $p_t(f(x))$ can be reformulated as {\begin{align}\sum_{i=1}^c{p_s(f(x)|y=i)p_s(y=i)}=\sum_{i=1}^c {p_t(f(x)|y=i)}p_t(y=i). \label{eq:ob}\end{align}} 

Suppose that the samples of class $i$, $x\in \mathcal{X}_i$, are completely distinguishable from the samples in the other classes in $f(x)$, with a decent $f(\cdot)$, $i.e.,$ {\begin{equation}
\begin{aligned}
    &p_s(f(x\in\mathcal{X}_i)|y=i)>0\Leftrightarrow p_s(f(x\in\mathcal{X}_i)|y\neq i)=0\\
    &p_t(f(x\in\mathcal{X}_i)|y=i)>0\Leftrightarrow p_t(f(x\in\mathcal{X}_i)|y\neq i)=0.  \label{eq:m1}
\end{aligned}\end{equation}}Therefore, advUDA is enforcing $p_s(f(x\in\mathcal{X}_i)|y=i)$ $p_s(y=i)$ $={p_t(f(x\in\mathcal{X}_i)|y=i)}p_t(y=i)$. Taking the integral of $x$ over $\mathcal{X}_i$ for both sides of the above equation, we have $p_s(y=i)=p_t(y=i)$. This deduction contradicts with the label shift setting $p_s(y)\neq p_t(y)$.

\begin{figure}[t]
\begin{center}
\includegraphics[width=1\linewidth]{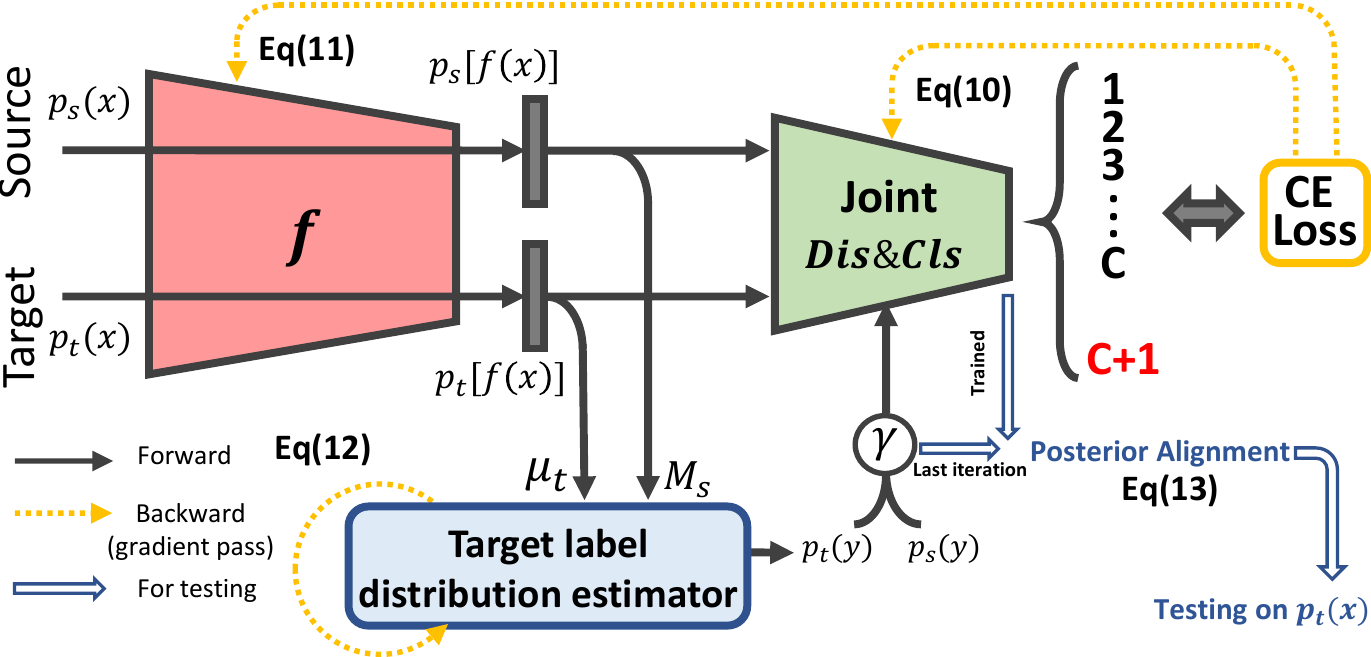}
\end{center} 
\caption{Illustration of our alternative training-based CLS framework, which consists of feature encoder $f$, joint $Dis$ and $Cls$, and $p_t(y)$ estimator. The posterior alignment is only applied at the testing stage. The overall objectives are shown in Eqs. (10-13), which are optimized iteratively. CE indicates cross-entropy.}
\label{fig:4} 
\end{figure}

Based on the above analysis, we have the conclusion that the label and conditional shifts are non-negligible in adversarial UDA. Since $p_t(y)$ is not available in UDA, we propose to align the class-dependent conditional distribution and estimate the label distribution, following a turn-based alternative strategy in each iteration, as shown in Fig. \ref{fig:4}.

\subsection{Conditional Shift Alignment}

To incorporate the knowledge of $p_t(y)$ to align $p_s(x|y)$ with $p_t(x|y)$, we introduce a class-wise balancing parameter $\gamma\in \mathbb{R}^{c}$. We expect that $\gamma_i{p_s(y=i)}={p_t(y=i)}$ and $\gamma^*$ denote the vector that makes this equation hold. Therefore, we propose to multiply $\gamma$ to the source domain term of the optimization objective in Eq.~(\ref{eq:ob}), $i.e.,$ $\sum_{i=1}^c{p_s(f(x)|y=i)p_s(y=i)\gamma_i}=\sum_{i=1}^c {p_t(f(x)|y=i)}p_t(y=i)$. Since $\gamma_i=\frac{{p_t(y=i)}}{{p_s(y=i)}}$, it can be written as $\sum_{i=1}^c{p_s(f(x)|y=i){\cancel{p_s(y=i)}}\frac{{p_t(y=i)}}{\cancel{p_s(y=i)}}}=\sum_{i=1}^c {p_t(f(x)|y=i)}p_t(y=i)$ and the estimated $p_t(y=i)$ is a constant in each iteration. Theoretically, with a precisely estimated $\gamma$, we can align the conditional distribution using adversarial UDA.

Besides, from an implementation perspective, it is important to make the discriminator be aware of classification. As discussed in \cite{saito2017maximum}, the three-player ($f, Cls, Dis$) game should fail to consider the relationship between the target samples and the task-specific decision boundary, when aligning distributions. Here, we simply adopt the jointly parameterized classifier and discriminator \cite{salimans2016improved,tran2019gotta} to incorporate the classification into the domain alignment. Next, we align $p_s(x|y)$ and $p_t(x|y)$ with the estimated $p_t(y)$, even though the up-to-date class-aware domain adaptation method can be added, following a plug-and-play fashion. 

We configure $c+1$ output units to represent the $c$ class and target domain, respectively. The score function ${\overline{C}}$ (joint $cls$ and $dis$) is defined on $\mathbb{R}^K\times\{1,\cdots,c+1\}$. Since the sum of the probability that $f(x)$ belongs to one of the $c$-class is not 1, we need to reparamiterize the conditional score: \begin{align}
  &{\overline{C}}(f(x),y|\mathcal{Y})=\frac{{\overline{C}}(f(x),y)}{1-{\overline{C}}(f(x),c+1)}\nonumber\\&\forall y\leq c; ~~~{\overline{C}}(f(x),c+1|\mathcal{Y})=0
\end{align}  Its optimization objective is given by 
\begin{align}
   ~_{Dis\&Cls}^{\rm{~~max}}~\{~&^\mathbb{~~E}_{x\sim p_s} {\rm log} {\overline{C}}(f(x),y|\mathcal{Y})\nonumber\\
   &+^\mathbb{~~E}_{x\sim p_s} {\rm log}(1-{\overline{C}}(f(x),c+1))\nonumber\\
   &+^\mathbb{~~E}_{x\sim p_t} {\rm log}({\overline{C}}(f(x),c+1))\}\label{8}
\end{align}  
\begin{align}
   ~~~~~~~~~~~~~_{~~f}^{\rm{max}}~\{~&^\mathbb{~~E}_{x\sim p_s} {\rm log} {\overline{C}}(f(x),y|\mathcal{Y})\nonumber\\
   &+\lambda^\mathbb{~~E}_{x\sim p_t} {\rm log}(1-{\overline{C}}(f(x),c+1))\}~~~~~~
\end{align}

The empirical effectiveness of the joint parameterization of $Cls$ and $Dis$ in semi-supervised learning has been shown in \cite{salimans2016improved}. \cite{tran2019gotta} give an intuitive explanation that all target examples yield the gradient of the same direction in the conventional three-player game. In this paper, we give a theoretical analysis from the domain adaptation theory perspective.

\vspace{+5pt}
\noindent\textbf{Proposition 1} The joint parameterization can lead to a better approximation of $\mathcal{L}_t(h)$ by $\mathcal{L}_s(h)$ in Theorem \textcolor{red}{1}. 
\vspace{+5pt}


This means that the performance in the target domain can be less affected by the domain shift. Please note that using only the joint parameterization cannot guarantee the conditional alignment, but can provide a powerful backbone for our conditional alignment scheme. We also give the detailed ablation study in our experiments.

In general, the expectation function in Eqs. (\textcolor{red}{8-9}) has the property $~^\mathbb{~~E}_{x\sim p_s(x)}[\cdot]\equiv$ $~^\mathbb{~~~~~~~~E}_{x\sim p_s(x|y)p_s(y)}[\cdot] \equiv \sum_{i=1}^c p_s(y=i)~^\mathbb{~~~~~~~~E}_{x\sim(p_s(x|y=i))}[\cdot]$, where $[\cdot]$ is a deterministic function of $x$ $w.r.t.$ $f$. Therefore, to take the target label distribution into account, we can make Eqs. (\textcolor{red}{8-9}) be aware of the estimated $\gamma$, by rewriting them as 
\begin{align}
   _{D\&C}^{\rm{~max}}\{&\sum_{i=1}^c\gamma_ip_s(y=i)^\mathbb{~~E}_{x\sim p_s(x|y=i)}{\rm log} {\overline{C}}(f(x),y|\mathcal{Y})~~~~~~~~~~\nonumber\\
   &+^\mathbb{~~E}_{x\sim p_s} {\rm log}(1-{\overline{C}}(f(x),c+1))\nonumber\\
   &+^\mathbb{~~E}_{x\sim p_t} {\rm log}({\overline{C}}(f(x),c+1))\} 
\end{align}
\begin{align}
   ~~~~_{~~f}^{\rm{max}}~\{&\sum_{i=1}^c \gamma_i p_s(y=i)~^\mathbb{~~E}_{x\sim p_s(x|y=i)} {\rm log} {\overline{C}}(f(x),y|\mathcal{Y})~~~~~~~~~~~~\nonumber\\&+\lambda^\mathbb{~~E}_{x\sim p_t} {\rm log}(1-{\overline{C}}(f(x),c+1))\}.
\end{align}

When $\gamma_i=1,\forall i\in\mathcal{Y}$, this alignment will degrade to conventional adversarial UDA. $\gamma_i$ can also be regarded as a class-level weight for each class in the source domain. This is essentially different from the sample-level weight in \cite{cao2019learning,zhang2018importance}. Actually, we multiply $\gamma_i$ to the $p_s(y=i)$ term, $i.e.,$ revising the source label distribution to the estimated target label distribution $\frac{{{p_t}(y=i)}}{\cancel{{p_s}(y=i)}}{\cancel{{p_s}(y=i)}}$.

\begin{table*}[!t]
\centering
\caption{Experimental results for the VisDA17-val setting. We use ResNet101 as our backbone except for \cite{pinheiro2018unsupervised,sankaranarayanan2018generate} and CLS:Res152.}
\resizebox{\linewidth}{!}{
\centering
\begin{tabular}{c|cccccccccccc|c}
\hline
Method         &  Aero & Bike   & Bus & Car & Horse & Knife & Motor   & Person   & Plant & Skateboard & Train  & Truck   & Mean \\ \hline\hline
Source-Res101 \cite{gholami2019taskdiscriminative} & 55.1 & 53.3 & 61.9 & 59.1 & 80.6 & 17.9 & 79.7 & 31.2 & 81.0 & 26.5 & 73.5 & 8.5 & 52.4 \\\hline

DANN (Baseline) \cite{ganin2016domain}  & 81.9 & 77.7 & 82.8 & 44.3 & 81.2 & 29.5 & 65.1 & 28.6 & 51.9 & 54.6 & 82.8 & 7.8 & 57.4 \\

MCD \cite{saito2017maximum} & 87.0 & 60.9 & \textbf{83.7} & 64.0 & 88.9 & 79.6 & 84.7 &  {76.9} &  {88.6} & 40.3 & 83.0 & 25.8 & 71.9 \\

ADR \cite{saito2017adversarial} & 87.8 & 79.5 & \textbf{83.7} & 65.3 & \textbf{92.3} & 61.8 &  {88.9} & 73.2 & 87.8 & 60.0 &  {85.5} & {32.3} & 74.8 \\  

DEV \cite{you2019toward} & 81.83& 53.48& 82.95& 71.62& 89.16 &72.03& 89.36& 75.73& \textbf{97.02} & 55.48 &71.19 &29.17& 72.42\\

TPN \cite{pan2019transferrable} &  93.7 &  85.1 &  69.2  & 81.6  & 93.5  & 61.9  & 89.3 &  81.4  & 93.5 &  81.6  & 84.5  & 49.9  & 80.4\\

CRST \cite{zou2019confidence} & 89.2 &79.6 &64.2 &57.8 &87.8 &79.6 &85.6 &75.9 &86.5 &85.1 &77.7 &68.5 &78.1\\

TDDA \cite{gholami2019taskdiscriminative} & 88.2 &78.5& 79.7& 71.1 &90.0 &81.6& 84.9 &72.3 &92.0& 52.6& 82.9& 18.4 &74.03\\

PANDA \cite{hu2020panda} &90.9 &50.5 &72.3& 82.7 &88.3 &88.3 &90.3& 79.8& 89.7 &79.2& 88.1 &39.4& 78.3\\

DMRL \cite{wu2020dual} & - &-& -&- &- &-& - &- &-& -& -& - &75.5\\

\hline

\textbf{CLS w/o Joint Dis+Cls} & 91.8$\pm$1.2 & 84.4$\pm$1.9 & 73.3$\pm$1.7 & 71.5$\pm$1.6 & 87.2$\pm$1.0 & 80.8$\pm$1.4 & 85.6$\pm$1.8 & 76.3$\pm$1.5 & 85.6$\pm$1.8 & 88.5$\pm$1.3 & 85.7$\pm$1.1 & 71.0$\pm$1.4 & 80.8$\pm$0.6 \\


\textbf{CLS w/o fade-in $p_t(y)$ } & 92.0$\pm$1.4 & 84.2$\pm$1.2 & 72.8$\pm$1.6 & 72.5$\pm$1.2 & 87.6$\pm$1.5 & {82.7$\pm$2.0} & 86.4$\pm$1.7 & 77.0$\pm$1.3 & 86.3$\pm$1.7 & 89.6$\pm$1.5 & 85.5$\pm$1.7 & 72.1$\pm$1.4 & 81.3$\pm$0.5 \\

\textbf{CLS} & 92.6$\pm$1.5 & 84.5$\pm$1.4 & 73.7$\pm$1.5 & 72.7$\pm$1.1 & 88.5$\pm$1.4 & 83.3$\pm$1.2 & 89.1$\pm$1.4 & 77.6$\pm$1.5 & 89.5$\pm$1.5 & 89.2$\pm$1.3 & 85.8$\pm$1.2 & 72.7$\pm$1.3 & 81.6$\pm$0.4\\
\hline
 
\textbf{CLS+TDDA w/o Joint Dis\&Cls} & 92.3$\pm$1.5 & 84.6$\pm$1.2 & 74.4$\pm$1.5 & 73.7$\pm$1.8 & 89.6$\pm$1.8& {82.5$\pm$1.9} & 88.4$\pm$1.5 & \textbf{82.8$\pm$1.5} & 89.1$\pm$1.6 & \textbf{91.9$\pm$1.5} & 85.6$\pm$1.6 & 73.5$\pm$1.3 & 82.2$\pm$0.5 \\


\textbf{CLS+TDDA} & \textbf{94.0$\pm$1.6} & \textbf{85.4$\pm$1.6} & 75.4$\pm$1.0 & \textbf{74.5$\pm$1.4} & 92.2$\pm$1.5 & \textbf{83.4$\pm$1.5} & 89.8$\pm$1.6 & \textbf{82.8$\pm$1.2 }& 90.8$\pm$1.2 & 91.8$\pm$1.4 & \textbf{87.5$\pm$1.2} & \textbf{74.1$\pm$0.9 }& \textbf{82.9$\pm$0.5}\\ 

\hline\hline

SimNet:Res152 \cite{pinheiro2018unsupervised} & \textbf{94.3} & 82.3 & 73.5 & 47.2 & 87.9 & 49.2 & 75.1 & 79.7 & 85.3 & 68.5 & 81.1 & 50.3 & 72.9 \\

GTA:Res152 \cite{sankaranarayanan2018generate}  
& - & - & - & - & - & - & - & - & - & - & - & - & 77.1\\\hline

\textbf{CLS:Res152 }& {94.2$\pm$1.3} & \textbf{86.8$\pm$1.2} & \textbf{74.7$\pm$1.0 }& \textbf{75.3$\pm$1.5 }& \textbf{90.5$\pm$0.6} & \textbf{88.6$\pm$1.6} & \textbf{89.2$\pm$1.6 }& \textbf{83.2$\pm$1.6 }& \textbf{91.9$\pm$1.3} & \textbf{92.6$\pm$1.8} & \textbf{86.1$\pm$1.4} & \textbf{80.9$\pm$1.4} & \textbf{83.8$\pm$0.5 }\\

\hline

\end{tabular}%
}

\label{table:visda17}
\end{table*}

\subsection{Target Label Distribution Estimation}

Assuming that the conditional distribution is aligned ($i.e.,$ $p_s(f(x)|y)=p_t(f(x)|y)$), we can estimate the target label distribution $p_t(y)$ using the marginal matching \cite{zhang2013domain}. Instead of working on raw data, in this work, we carry out our analyses in the feature space. Note that $p_s(y)$ can be obtained, by counting each labeled sample. 

Targeting this, the mean matching \cite{gretton2009covariate} is a simple yet efficient method. The target label proportion $p_t(y)$ can be estimated, by minimizing the loss function {{\begin{align}
    \mathcal{L}_M(p_t(y))=||M_sp_t(y)-\mu_t||_2^2,
\end{align}}}where $M_s$ is the vector of $[\mu_s(f(x)|y=1), \mu_s(f(x)|y=2), \cdots, \mu_s(f(x)|y=c))]$, i.e., the empirical sample means from the source domain, and $\mu_t$ is the encoded feature mean of the target. Note that the above loss is equivalent to a standard linearly constrained quadratic problem, resulting in the estimated target label proportions $p_t(y)$. Typically, this can be solved by gradient descent in each minibatch.

Suppose that the conditional shift can be aligned, then the target label distribution estimation scheme is asymptotically consistent, which can be expressed below.

\vspace{+5pt}
\noindent\textbf{Theorem 3} Assuming $p_s(f(x)|y)=p_t(f(x)|y)$, the variance in the feature space is not infinite, and all of the label proportions are not zero. As the number of training and testing samples becomes infinity, the estimated ${p}_t(y)$ is asymptotically consistent for real ${p}_t(y)$, if $M_s^{\rm T}M_s$ is invertible.\vspace{+5pt}

Here the superscript ${\rm T}$ represents transpose. While this theorem only considers a single source domain, it can be easily extended to multiple source domains.

We note that the L2 loss used in $\mathcal{L}_M(p_t(y))$ can also be replaced by the other distribution loss, $e.g., f$-divergence, Wasserstein loss, or MMD to form the distribution matching. In some cases, the model can be benefited, by matching more than the first moment. 

Considering that the segmentation task is to perform pixel-wise classification, the label distribution is essentially the proportion of the class-wise pixel numbers in a dataset.

\begin{table*}[!t]
\centering
\caption{Experimental results for GTA5 to Cityscapes.}
\resizebox{1\linewidth}{!}{
\centering
\begin{tabular}{c|c|ccccccccccccccccccc|c}
\hline
Method         & Base Net          & Road & SW   & Build & Wall & Fence & Pole & TL   & TS   & Veg. & Terrain & Sky  & PR   & Rider & Car  & Truck & Bus  & Train & Motor & Bike & mIoU \\ \hline\hline
Source     & DRN26 & 42.7 & 26.3 & 51.7  & 5.5  & 6.8   & 13.8 & 23.6 & 6.9  & 75.5 & 11.5    & 36.8 & 49.3 & 0.9   & 46.7 & 3.4   & 5.0  & 0.0   & 5.0   & 1.4  & 21.7 \\
CyCADA \cite{hoffman2018cycada}    &   & 79.1 & 33.1 & 77.9  & 23.4 & 17.3  & 32.1 & 33.3 & 31.8 & 81.5 & 26.7 & 69.0 & 62.8 & 14.7  & 74.5 & 20.9  & 25.6 & 6.9   & 18.8  & 20.4 & 39.5 \\ \hline
Source     & DRN105          &  36.4 & 14.2 & 67.4 & 16.4 & 12.0 & 20.1 & 8.7 & 0.7 & 69.8 & 13.3 & 56.9 & 37.0 & 0.4 & 53.6 & 10.6 & 3.2 & 0.2 & 0.9 & 0.0 & 22.2 \\
MCD \cite{saito2017maximum}  &   &90.3 & 31.0 & 78.5 & 19.7 & 17.3 & 28.6 & 30.9 & 16.1 & 83.7 & 30.0 & 69.1 & 58.5 & 19.6 & 81.5 & 23.8 & 30.0 & 5.7 & 25.7 & 14.3 & 39.7 \\ \hline
Source & PSPNet & 69.9 & 22.3 & 75.6 & 15.8 & 20.1 & 18.8 & 28.2 & 17.1 & 75.6 & 8.00 & 73.5 & 55.0 & 2.9 & 66.9 &  {34.4} & 30.8 & 0.0 & 18.4 & 0.0 & 33.3 \\
DCAN \cite{wu2018dcan} & & 85.0 & 30.8 & 81.3 & 25.8 & 21.2 & 22.2 & 25.4 & 26.6 & 83.4 & {36.7} & 76.2 & 58.9 & 24.9 & 80.7 & 29.5 & 42.9 & 2.50 & 26.9 & 11.6 & 41.7 \\ \hline

Source&ResNet-101 &75.8 &16.8 &77.2 &12.5 &21.0&25.5 &30.1 &20.1 &81.3 &24.6 &70.3& 53.8& 26.4 &49.9 &17.2 &25.9& 6.5& 25.3 &36.0 &36.6\\

DPR \cite{tsai2019domain}& &92.3 &51.9 &82.1 &29.2& 25.1& 24.5& 33.8&  {33.0} &82.4& 32.8& {82.2}& 58.6& 27.2& 84.3 &33.4& {46.3} &2.2 &29.5 &32.3& 46.5\\\hline

Source&ResNet-101 &73.8& 16.0& 66.3& 12.8 &22.3& 29.0& 30.3 &10.2 &77.7 &19.0 &50.8 &55.2& 20.4& 73.6& 28.3 &25.6 &0.1& 27.5 &12.1 &34.2\\

PyCDA \cite{lian2019constructing} && 90.5&  36.3&  84.4&  32.4 &  {28.7} & 34.6&  36.4 & 31.5&  86.8&  37.9&  78.5 & 62.3 & 21.5&  85.6&  27.9 & 34.8&  18.0 & 22.9&  \textbf{49.3} & 47.4\\\hline

Source & ResNet101 & 71.3 & 19.2 & 69.1 & 18.4 & 10.0 & 35.7 & 27.3 &  6.8 & 79.6 & 24.8 & 72.1 & 57.6 & 19.5 & 55.5 & 15.5 & 15.1 & 11.7 & 21.1 & 12.0 & 33.8 \\

CRST    \cite{zou2019confidence}     &                   & 89.0 & 51.2 & 79.4 & 31.7 & 19.1 & 38.5 & 34.1 & 20.4 & 84.7 & 35.4 & {76.8} & 61.3 & {30.2} & 80.7 & 27.4 & 39.4 & 10.2 & 32.2 & {43.3} & {46.6} \\\hline\hline

\hline

Source&ResNet-101 &75.8 &16.8 &77.2 &12.5 &21.0&25.5 &30.1 &20.1 &81.3 &24.6 &70.3& 53.8& 26.4 &49.9 &17.2 &25.9& 6.5& 25.3 &36.0 &36.6\\

APODA \cite{yang2020adversarial}&   & 85.6 &32.8& 79.0& 29.5& 25.5& 26.8& 34.6 &19.9& 83.7& 40.6& 77.9 &59.2& 28.3& 84.6& 34.6 &49.2& 8.0 &32.6& 39.6 &45.9\\\hline

\textbf{CLS+\cite{yang2020adversarial}}    &    & \textbf{94.7} & 59.2 & 82.6 & 31.1 & 26.7 & 42.3 & 38.5 & 29.2 & 86.3 & 39.0 & 78.9 & \textbf{67.6} & 36.1 & 85.8 & 28.3 & 45.6 & 21.8 & {36.8} & 46.4 & 51.2 \\

\textbf{CLS+\cite{yang2020adversarial} w/o Joint Dis+Cls}  &   & 94.1 & 54.9 & 81.6 & 32.6 & 22.0 & 41.4 & 36.0 & 20.8 & {87.6} & 36.9 & 77.4 &{64.3}& 32.4 & 84.0 & 23.5 & 39.6 & 25.2 & 33.2 & 40.4 & 50.8 \\\hline\hline


Source&ResNet-101 &75.8 &16.8 &77.2 &12.5 &21.0&25.5 &30.1 &20.1 &81.3 &24.6 &70.3& 53.8& 26.4 &49.9 &17.2 &25.9& 6.5& 25.3 &36.0 &36.6\\

IAST-MST \cite{ke2020instance}&&94.1& 58.8& 85.4 &\textbf{39.7}& \textbf{29.2}& 25.1& \textbf{43.1}& \textbf{34.2}& 84.8&34.6 &\textbf{88.7}& 62.7& 30.3& \textbf{87.6}& \textbf{42.3}& \textbf{50.3}& 24.7 &35.2& 40.2& 52.2\\\hline

\textbf{CLS+\cite{ke2020instance}} &         & \textbf{94.7} & 60.1 & \textbf{85.6} &  {39.5} & 24.4 & \textbf{44.1} &  {39.5} & 20.6 & \textbf{88.7} & 38.7 & 80.3 & {67.2}
& 35.1 &  {86.5} & 37.0 & 45.4 & \textbf{39.0} & \textbf{37.9} & 46.2 & \textbf{53.0}  \\

\textbf{CLS+\cite{ke2020instance} w/o Joint Dis+Cls  }    &   & 94.5 & \textbf{58.6} & 84.9 & 32.2 & 26.5 & 42.4 & 39.3 & 28.4 & 88.6 & \textbf{42.3} & 79.5 & 67.4 & \textbf{36.5} & 83.3 & 34.8 & 43.2 & 24.6 & {36.5} & 48.3 & 52.5\\  \hline

\end{tabular}%
}

\label{table:gtacity}
\end{table*}

\subsection{Alternative Optimization with Fade-in $p_t(y)$} 

In summary, we have three to-be learned players in the game at our training stage, i.e., feature extractor $f(\cdot)$, joint $Dis\&Cls$, and a $p_t(y)$ estimator. In each iteration, we update them following an alternative optimization scheme:

\noindent{\textbf{[Infer]}}: Fix $f(\cdot)$ and update the target label distribution estimator with $\frac{ \partial\mathcal{L}_M(p_t(y))}{ \partial p_t(y)}$.

\noindent{\textbf{[Align]}}: Fix $p_t(y)$ estimator first and update the feature extractor and joint $Dis\&Cls$ following the GAN's protocol according to Eq. (\textcolor{red}{10}) and Eq. (\textcolor{red}{11}), respectively.

Considering that it is difficult to precisely estimate $p_t(y)$ at the start of the training, we simply initialize it as uniform distribution, i.e., $p_t(y=i)=\frac{1}{c}$ and construct the target label distribution with $\frac{1}{1+\alpha}\left\{p_t(y)+\alpha p_s(y)\right\}$ in each round. $\alpha$ is decreased from 1 to 0 gradually at the training stage, i.e., $\alpha=\frac{1}{1+N}$ for epoch number $N\leq$5 and $\alpha=0$ for $N>$5.

\subsection{Posterior Alignment} 

Finally, we obtain the target domain classifier $p_t(y=i|f(x))$ in testing. Suppose that we have successfully aligned the conditional shift and precisely estimated the target label distribution at the training stage, following \cite{chan2005word}, the posterior in the target domain can be $p_t(y=i|f(x))=\frac{\gamma_i^*p_s(y=i|f(x))}{\sum_{j=1}^cp_s(y=j|f(x))}
$. Since we cannot practically know the exact $\gamma*$, we choose the estimated $p_t(y)$ to calculate $\gamma$ in each iteration, and approximate the posterior by {\begin{align} p_t(y=i|f(x))\leftarrow\frac{\gamma_i p_s(y=i|f(x))}{|\gamma|_1\sum_{j=1}^cp_s(y=j|f(x))}.\end{align}}

\section{Experimental Results}

We provide comprehensive evaluations of our framework under the conditional and label shift (CLS) on image classification and semantic segmentation (partial) UDA tasks.

\begin{table*}[]
\centering\caption{Experimental results for partial unsupervised domain adaptation on Office-31 dataset (ResNet-50).}
\resizebox{0.85\linewidth}{!}{%
\begin{tabular}{c|ccccccc}

\hline
Method & A31$\rightarrow$W10 & D31$\rightarrow$W10 & W31$\rightarrow$D10 & A31$\rightarrow$D10 & D31$\rightarrow$A10 & W31$\rightarrow$A10 & Mean   \\ \hline

ResNet-50 \cite{he2016deep} & 75.59$\pm$1.09 & 96.27$\pm$0.85 & 98.09$\pm$0.74 & 83.44$\pm$1.12 & 83.92$\pm$0.95 & 84.97$\pm$0.86 & 87.05$\pm$0.94  \\\hline

DAN \cite{long2015learning}  & 59.32$\pm$0.49 & 73.90$\pm$0.38 & 90.45$\pm$0.36 & 61.78$\pm$0.56 & 74.95$\pm$0.67 & 67.64$\pm$0.29 & 71.34$\pm$0.46  \\

DANN \cite{ganin2016domain} & 73.56$\pm$0.15 & 96.27$\pm$0.26 & 98.73$\pm$0.20 & 81.53$\pm$0.23 & 82.78$\pm$0.18 & 86.12$\pm$0.15 & 86.50$\pm$0.20   \\

ADDA \cite{tzeng2017adversarial}  & 75.67$\pm$0.17 & 95.38$\pm$0.23 & 99.85$\pm$0.12 & 83.41$\pm$0.17 & 83.62$\pm$0.14 & 84.25$\pm$0.13 & 87.03$\pm$0.16  \\

IWAN \cite{zhang2018importance}  & 89.15$\pm$0.37 & 99.32$\pm$0.32 & 99.36$\pm$0.24 & 90.45$\pm$0.36 & 95.62$\pm$0.29 & 94.26$\pm$0.25 & 94.69$\pm$0.31  \\

SAN \cite{cao2018partialb} & 93.90$\pm$0.45 & 99.32$\pm$0.52 & 99.36$\pm$0.12 & 94.27$\pm$0.28 & 94.15$\pm$0.36 & 88.73$\pm$0.44 & 94.96$\pm$0.36  \\ 

PADA  \cite{cao2018partiala} & 86.54$\pm$0.31 & 99.32$\pm$0.45 & \textbf{100.00$\pm$0.00} & 82.17$\pm$0.37 & 92.69$\pm$0.29 & {95.41$\pm$0.33} & 92.69$\pm$0.29  \\

ETN \cite{cao2019learning} & 94.52$\pm$0.20 & \textbf{100.00$\pm$0.00} & \textbf{100.00$\pm$0.00} & \textbf{95.03$\pm$0.22} & 96.21$\pm$0.27 & 94.64$\pm$0.24 & 96.73$\pm$0.16 \\

BAUS \cite{liang2020balanced}& 98.98$\pm$0.28&  \textbf{100.0$\pm$0.00}& 98.73$\pm$0.00& \textbf{99.36$\pm$0.00}& 94.82$\pm$0.05& 94.99$\pm$0.08 & 97.81  \\

\hline\hline

\textbf{CLS }& \textbf{99.64$\pm$0.28} & \textbf{100.00$\pm$0.00} & \textbf{100.00$\pm$0.00 }& 97.26$\pm$0.30 &   {97.94$\pm$0.23} &  \textbf{98.32$\pm$0.34} & \textbf{98.24$\pm$0.15}  \\

\textbf{CLS w/o Joint Dis+Cls} & 98.91$\pm$0.35 & 99.62$\pm$0.43 &  \textbf{100.00$\pm$0.00} & 96.36$\pm$0.27 & \textbf{98.05$\pm$0.25} & 97.75$\pm$0.42 & 97.94$\pm$0.32 \\ 


\hline
\end{tabular}%
}

\label{tabel:office22}
\end{table*}

\subsection{UDA for Image Classification}

We first consider the challenging VisDA17 benchmark \cite{peng2018visda}. It is a 12-class UDA classification problem. We follow the evaluation protocols as in \cite{sankaranarayanan2018generate,saito2017adversarial}, where a source domain incorporates $152,409$ synthetic images and a target domain has $55,400$ real-world images.

For fair comparisons, we adopt the standard ResNet101 \cite{he2016deep} as our backbones for VisDA17. We use ImageNet \cite{deng2009imagenet} for pre-training, and fine-tune the network in the source domain with SGD and set $lr=1\times10^{-3}$. The batch size is set to 32 consistently. We simply set $\lambda=1$ in all the experiments. 


We show the results on VisDA17 in Table \ref{table:visda17} in terms of the per-class accuracy and mean accuracy. To report the standard deviation of each method, we independently run five times. The CLS w/o Joint Dis+Cls, and CLS w/o fade-in $p_t(y)$ denote the CLS with separate discriminator and classifier and without fade-in $p_t(y)$, respectively.

DANN \cite{ganin2016domain} is a pioneer of vanilla adversarial UDA, and its performance on VisDA is reported in \cite{gholami2019taskdiscriminative}. Benefited by the conditional and label alignments, CLS outperforms DANN by a large margin and the recent adversarial UDA method, TDDA, significantly, while the joint parameterization further boosts its performance. Although the full training of CLS takes 1.5 times more training time than DANN, it only requires 2$\times$ less training time to obtain comparable results than DANN. Moreover, CLS has the same inference speed as conventional adversarial UDA methods at its testing stage. Note that the CDAN \cite{long2018conditional} method is ill-posed as it only considers the conditional shift in UDA (without $p_t(y)$), and a direct comparison with that can be helpful. However, its relatively old backbone (ResNet50) is not adopted by recent state-of-the-art approaches (SOTA). Our CLS with ResNet50 can also achieve a mean accuracy of 76.2\% on VisDA17, which is a significant improvement over CDAN (70\%). Besides, Our CLS outperforms DEV on VisDA17 (all with ResNet101), and DEV outperforms CDAN on Office31 (all with ResNet50).

\begin{figure}[t]
\begin{center}\caption{The KL-divergence between the predicted $p_t(y)$ and the ground truth $p_t(y)$ at the testing stage. Left: VisDA17 task, right: GTA5 to Cityscapes task. The smaller the value is the more accurate the estimation is.}
\includegraphics[width=1\linewidth]{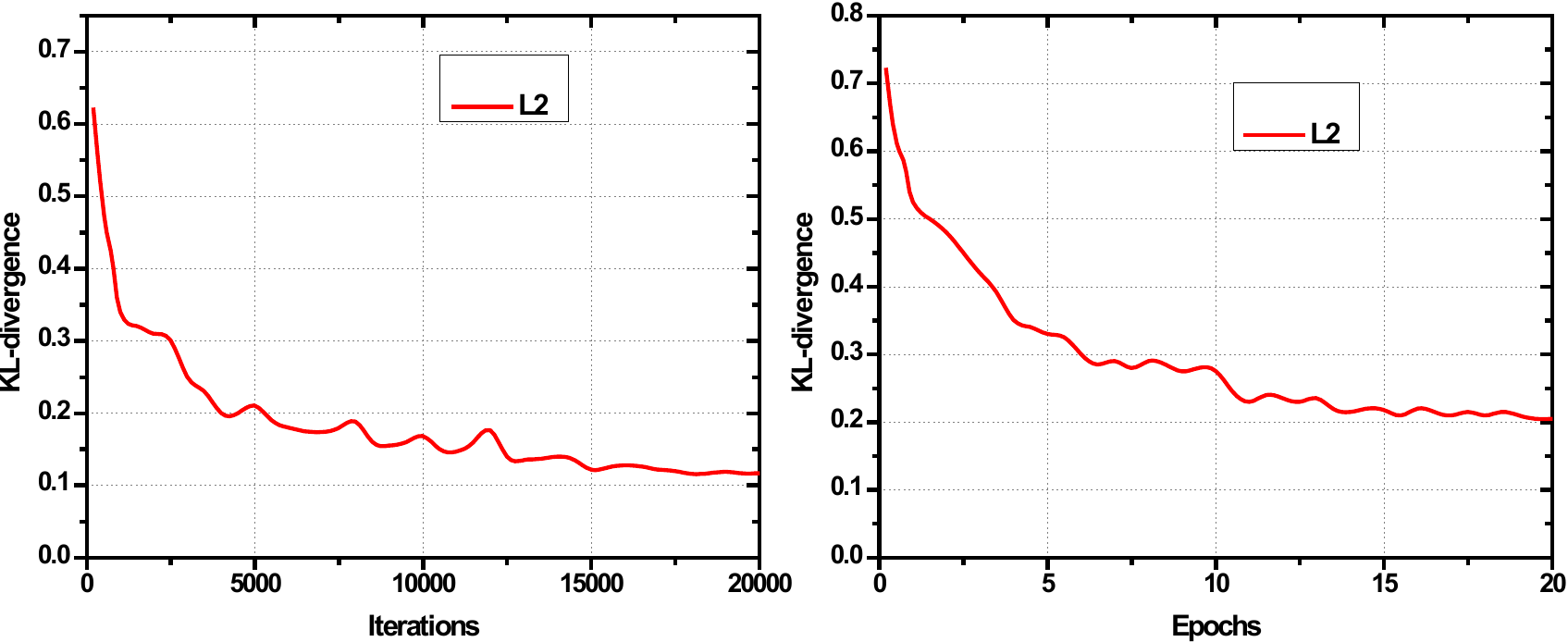}
\end{center}

\label{fig:6}
\end{figure}

Moreover, our CLS can be orthogonal with other advanced adversarial UDA methods. For example, the regularization losses used in TDDA \cite{gholami2019taskdiscriminative} can be simply add-on and further improve the performance. To the best of our knowledge, all CLSs, including CLS w/o Joint Dis+Cls, and CLS w/o fade-in $p_t(y)$ achieve the STOA of adversarial UDA in VisDA 17 with ResNet 101 as the backbone.

More appealingly, CLS achieves on par or better performance than the recent adversarial training \cite{wu2020dual,gholami2019taskdiscriminative}, self-training \cite{zou2019confidence}, dropout \cite{saito2017adversarial}, and moment matching methods \cite{pan2019transferrable,hu2020panda}, demonstrating the potential of adversarial learning in UDA. The more powerful backbones have also been applied and shown better results \cite{pinheiro2018unsupervised,sankaranarayanan2018generate}. The CLS with the ResNet152 backbone outperforms both CLS with the ResNet 101 backbone and \cite{pinheiro2018unsupervised,sankaranarayanan2018generate} significantly.

\subsection{UDA for Semantic Segmentation} Then, we investigate a more challenging segmentation adaptation case that is from synthetic to real images. We select the GTA5 \cite{richter2016playing} to Cityscapes task, which has 19 shared segmentation labels. There are more than 24,000 labeled game engine rendered images in the GTA5 dataset. According to the evaluation protocol in previous works \cite{yang2020adversarial}, the whole GTA5 dataset is used as the labeled source domain, and the training set of Cityscape with 2,975 images is used as the unlabeled target domain training data. During testing, we also report the performance in the testing set of Cityscapes with 500 images.

For fair comparisons with the other methods, we adopt the ResNet-101 as our backbone network for the GTA5 to Cityscapes task. Note that PSPNET \cite{zhao2017pyramid} and Wide ResNet38 are the stronger basenet than ResNet101. We compare CLSs with the other methods in Table \ref{table:gtacity}. Based on the adversarial training scheme of \cite{yang2020adversarial}, CLS outperforms \cite{yang2020adversarial}, by more than 4\% w.r.t. the mean IoUs. 



When compared with other state-of-the-art adversarial UDAs, CLS methods lead to a new state-of-the-art performance, even compared with a generative pixel-level domain adaptation method or adversarial and self-training combination methods, which are the relatively complex algorithm in terms of both architecture and objectives. We note that the additional self-training UDA can also be applied to our CLS, following the combination scheme in \cite{ke2020instance}, yielding 53.0 mIoU. This clearly demonstrates the generality of our CLS. Introducing our CLS to AdaptSegNet can significantly improve the segmentation performance. To investigate the target label estimation performance, we plot the KL-divergence between the predicted $p_t(y)$ and the ground truth $p_t(y)$ at the testing stage of both classification and segmentation tasks as in Fig. \ref{fig:6}.

Consistently with the classification, our CLS performs better, or on par with self-training-based methods \cite{zou2019confidence,lian2019constructing}, indicating the adversarial learning can still be a powerful methodology in UDA. Of note, the moment matching is usually not well scalable to segmentation.


\begin{table*}[!t]
\centering\caption{Experimental results for partial UDA on Cityscapes (merged 14-class) $\rightarrow$ NTHU (10-class). *Our reproduced results.}
\resizebox{0.8\linewidth}{!}{%
\begin{tabular}{c|c|cccccccccc|c}
\hline
Method         & Base Net    & Road & SW   & Build  & TL   & TS   & Veg.  & Sky  & PR  & Car  & Bus    & mIoU \\ \hline

Source &Dilation-Frontend \cite{yu2015multi}&80.3 &24.8 &69.3 &8.2 &5.5 &70.4 &\textbf{76.5} &40.2 &55.4 &18.7  &38.8\\

GCAA \cite{Tsai_adaptseg_2018}* & &81.0 &34.2& 69.6& 11.8 &13.4& 74.5& 63.1 &42.4& 60.1 &20.5 & 41.5\\

CBST \cite{Zou_2018_ECCV}* & &82.6 &35.1& \textbf{77.4}& 6.6 &28.5& \textbf{80.2}& 75.7 &\textbf{49.6}& 69.8 &4.8 & 45.4\\

\textbf{CLS} & &\textbf{84.6 }&\textbf{38.7}& 75.4& \textbf{19.3} &\textbf{32.6}& 77.4& 69.6 &49.3 & \textbf{70.2} &30.5 & \textbf{49.3}\\\hline\hline

Source &ResNet-101 & 83.2 &36.0& 70.6& 13.5& 12.8& 75.3& 61.2& 43.5& 63.8& 20.4 & 42.0\\

GCAA \cite{Tsai_adaptseg_2018}* & &81.6 &27.1& 73.5 &16.4 &23.8 &80.2 &87.5 &52.3  &68.1 &19.3 & 47.6\\

CBST \cite{Zou_2018_ECCV}* &  &84.5 &34.2& 79.6& 10.5 &29.1& 82.2& \textbf{79.4}&\textbf{54.7}& 74.0 &5.7& 48.2\\

\textbf{CLS} & &\textbf{88.1} &\textbf{42.3}& \textbf{83.6}& \textbf{26.7} &\textbf{37.5}& \textbf{87.3}& 64.6 &45.4& \textbf{79.2} &\textbf{24.5} & \textbf{50.7}\\\hline

\end{tabular}%
}

\label{table:gtacity2x}
\end{table*}

\begin{table}[]
\centering\caption{Experimental results for partial unsupervised domain adaptation on ImageNet-Caltech dataset (ResNet-50).}
\resizebox{1\linewidth}{!}{%
\begin{tabular}{c|ccc}

\hline

Method   &Img1000$\rightarrow$Cal84 & Cal256$\rightarrow$Img84 & Avg \\ \hline

ResNet-50 \cite{he2016deep}   & 69.69$\pm$0.78 & 71.29$\pm$0.74 & 70.49$\pm$0.76 \\\hline

DAN \cite{long2015learning}    & 71.30$\pm$0.46 & 60.13$\pm$0.50 & 65.72$\pm$0.48\\

DANN \cite{ganin2016domain}  & 70.80$\pm$0.66 & 67.71$\pm$0.76 & 69.23$\pm$0.71 \\

ADDA \cite{tzeng2017adversarial}   & 71.82$\pm$0.45 & 69.32$\pm$0.41 & 70.57$\pm$0.43 \\

IWAN \cite{zhang2018importance}  & 78.06$\pm$0.40 & 73.33$\pm$0.46 & 75.70$\pm$0.43 \\

SAN \cite{cao2018partialb}   & 77.75$\pm$0.36 & 75.26$\pm$0.42 & 76.51$\pm$0.39 \\ 

PADA  \cite{cao2018partiala}  & 75.03$\pm$0.36 & 70.48$\pm$0.44 & 72.76$\pm$0.40 \\

ETN \cite{cao2019learning}  & 83.23$\pm$0.24 & 74.93$\pm$0.28 & 79.08$\pm$0.26 \\

BAUS \cite{liang2020balanced} & 84.00$\pm$0.15& 83.35$\pm$0.28 &83.68\\

\hline\hline

\textbf{CLS }  & \textbf{85.13$\pm$0.35} & \textbf{84.47$\pm$0.42} & \textbf{84.76$\pm$0.39} \\

\textbf{CLS w/o Joint Dis+Cls}  &  83.05$\pm$0.37 & 84.14$\pm$0.45 &83.52$\pm$0.41 \\ 


\hline
\end{tabular}%
}

\label{tabel:office22x}
\end{table}


\subsection{Partial UDA for Image Classification}
To demonstrate it generally, we also apply our method to partial UDA task, i.e., a specific label shift setting, although CLS is developed for a more general problem. We note that partial UDA is not the extreme of label shift, when we use KL-divergence as the measurement.

Office-31 dataset \cite{saenko2010adapting} is the $de facto$ standard benchmark for domain adaptation. It contains 4,652 images with 31 classes from three domains: Amazon (A), Webcam (W) and DSLR (D), each containing $2,817$, $795$ and $498$ images respectively.

Following the experimental setting in \cite{cao2019learning,cao2018partiala}, we select 10 classes shared by Office-31 and Caltech-256 to form a new target domain. Note that there are 31 classes in the source domain and 10 classes in the target domain, and 21 classes have zero probability in the target label distribution.

We compare the performances of different methods on Office-31 with the same base net ResNet-50 in Table \ref{tabel:office22} left. All CLSs achieve better or comparable performance, compared with the state-of-the-art methods. This may be benefited, by also explicitly involving the conditional shift alignment along with addressing the label shift. The joint parameterization can consistently help the framework to align conditional and label distributions, respectively.

Besides, ImageNet-Caltech is a large scale dataset, which consists of ImageNet1K \cite{russakovsky2015imagenet} and Caltech \cite{griffin2007caltech} datasets. Since they share 84 classes, we typically use 1,000 categories ImageNet data as a source domain, while using 84 categories Caltech data as a target domain, or choose 256 categories Caltech as a source domain, while using 84 categories ImageNet as a target domain. Considering that the networks are usually trained on ImageNet, we follow the conventional partial UDA methods \cite{cao2019learning,cao2018partiala} for Caltech(256)$\rightarrow$ImageNet(84) adaptation, which uses the validation set of ImageNet as a target domain.

We give the comparison results in Table \ref{tabel:office22x} right. It is worth noticing that without considering the conditional and label distribution shift, DAN, DANN, and ADDA obtain worse performance than the ResNet-50 backbone in many tasks. To avoid the negative transfer in partial UDA setting, \cite{cao2019learning} propose a weighting mechanism to quantify the transferability of each source example. CSL outperforms or on par with the single example-based weighting methods \cite{cao2019learning} on most of the tasks, showing the efficiency of the class-based balancing scheme. Of note, the to-be-learned weights are much fewer than example-based weighting.

\subsection{Partial UDA for Semantic Segmentation}

To investigate a more challenging partial UDA scenario, we propose a new task for semantic segmentation. The Cityscapes to NTHU is a cross-city task with real-world images \cite{chen2017no}. We choose the Tokyo subset of the NTHU.

The NTHU \cite{chen2017no} has 13 segmentation classes shared with the Cityscapes dataset. Following the standard protocol in NTHU, the pole, fence, and wall are considered as a building. In addition, the truck is regarded as the car class, and the terrain is regarded as the vegetation class. According to the previous works \cite{chen2017no,Zou_2018_ECCV}, the 100 images in each city are separated to ten folds for the leave-one-out cross-validation. NTHU does not have the category of the train. To construct a more uneven target label distribution, we remove all of the images with the motor, bike, and rider classes in NTHU. Since the fact that motors are prohibited in many cities of China, and bikes are not allowed in highway can be used as prior knowledge.

As shown in Table \ref{table:gtacity2x}, our CLS works well with different popular backbones $i.e.,$ Dilation-Frontend Convolution as in \cite{Tsai_adaptseg_2018} and ResNet101 as in \cite{Zou_2018_ECCV}. We note that the example weighted adversarial partial UDA \cite{cao2019learning} and moment matching-based methods usually cannot be well scalable to the pixel-level classification-based segmentation task. CLS can be a strong baseline of partial UDA for semantic segmentation.

\section{Conclusion} 

In this paper, we tackle the problem of both conditional and label shifts in the context of adversarial UDA. We carry out a thoroughly theoretical analysis to demonstrate that it gives a more realistic and even minimal assumption in many tasks. Then, we propose to align these two shifts in an alternative optimization with a fade-in $p_t(y)$ framework which consists of three to-be learned players, i.e., feature extractor, joint parameterized discriminator and classifier, and target label distribution estimator. The merits of joint parameterization are also investigated. After the training, the posterior alignment is applied for real-world deployment. Extensive experiments on popular (partial) UDA benchmarks verify our analysis and revoke the adversarial learning, which can still be effective as the recent fast developed self-training, dropout or moment matching methods. 


\section*{Acknowledgments}
This work was supported in part by the Hong Kong GRF 152202/14E, PolyU Central Research Grant G-YBJW, and Jiangsu NSF (BK20200238).

{\small
\bibliographystyle{ieee_fullname}
\bibliography{main}
}

\end{document}